\documentclass[10pt,twocolumn,letterpaper]{article}

\usepackage[pagenumbers]{cvpr} 
\usepackage{amsmath}
\usepackage{amssymb}
\usepackage{bm}
\usepackage{algorithm}
\usepackage{algpseudocode}
\usepackage{subcaption}
\usepackage{varwidth}
\usepackage{scalerel}
\usepackage[accsupp]{axessibility}

\usepackage[dvipsnames]{xcolor}

\definecolor{cvprblue}{rgb}{0.21,0.49,0.74}
\usepackage[pagebackref,breaklinks,colorlinks,citecolor=cvprblue]{hyperref}

\def\paperID{16} 
\def\confName{CVPR}
\def\confYear{2024}

\title{Lacunarity Pooling Layers for Plant Image Classification using Texture Analysis}

\author{Akshatha Mohan and Joshua Peeples\\
Department of Electrical and Computer Engineering, Texas A\&M University\\
College Station, TX, USA\\
{\tt\small \{akshatha.mohan, jpeeples\}@tamu.edu}
}

\begin{document}
\maketitle
\begin{abstract}
Pooling layers (e.g., max and average) may overlook important information encoded in the spatial arrangement of pixel intensity and/or feature values. We propose a novel lacunarity pooling layer that aims to capture the spatial heterogeneity of the feature maps by evaluating the variability within local windows. The layer operates at multiple scales, allowing the network to adaptively learn hierarchical features. The lacunarity pooling layer can be seamlessly integrated into any artificial neural network architecture. Experimental results demonstrate the layer’s effectiveness in capturing intricate spatial patterns, leading to improved feature extraction capabilities. The proposed approach holds promise in various domains, especially in agricultural image analysis tasks. This work contributes to the evolving landscape of artificial neural network architectures by introducing a novel pooling layer that enriches the representation of spatial features. Our code is publicly available. \footnote
{\url{https://github.com/Advanced-Vision-and-Learning-Lab/2024_V4A_Lacunarity_Pooling_Layer}}
 
\end{abstract}    
\section{Introduction}
\label{sec:intro}


In computer vision, the common assumption of uniform pixel intensity in images fails to capture the complexity inherent in real-world images \cite{textureintro}. Visual textures that are easily understood by humans introduce intricate patterns, because of surface roughness and/or color disparities \cite{colourtexture}. These natural surfaces exhibit spatial distribution of local image patterns can exhibit statistical self-similarities. The notion of self-similarity is that an object is composed of sub-units and sub-sub-units on multiple levels that (statistically) resemble the structure of the whole object \cite{hutchinson1981fractals}.

\begin{figure}[t]
    \centering
    \setlength{\fboxsep}{0pt} 
    \begin{subfigure}[b]{0.3\linewidth} 
        \centering
        \includegraphics[width=\linewidth]{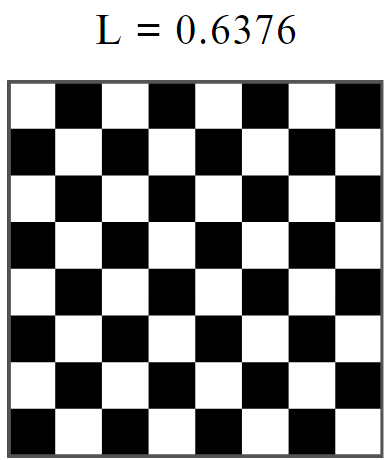}
        \caption{Low}
        \label{fig:image1}
    \end{subfigure}\hfill 
    \begin{subfigure}[b]{0.3\linewidth} 
        \centering
        \includegraphics[width=\linewidth]{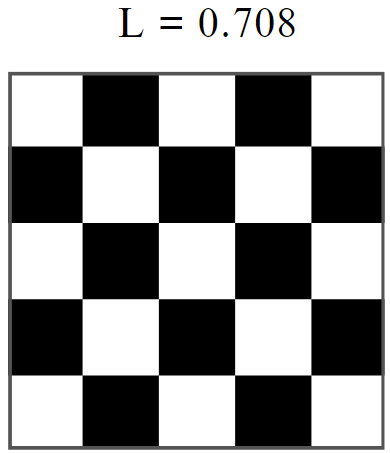}
        \caption{Medium}
        \label{fig:image2}
    \end{subfigure}\hfill 
    \begin{subfigure}[b]{0.3\linewidth}
        \centering
        \includegraphics[width=\linewidth]{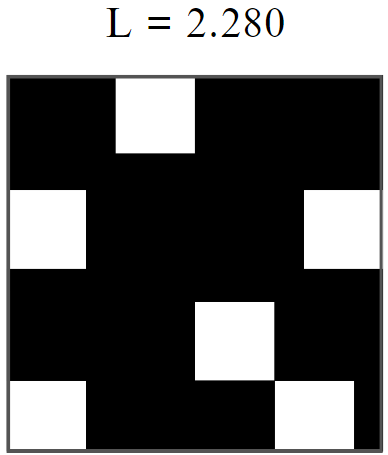}
        \caption{High}
        \label{fig:image3}
    \end{subfigure}\hfill
\caption{Fractal patterns exhibit an increase in lacunarity from left to right, indicating a rise in irregular gaps towards the right.}
\label{fig:lacunarityexample}
\end{figure}

For Agricultural image analysis, understanding the intricate textures and spatial patterns of plant images plays a pivotal role in tasks such as disease detection, crop monitoring, and yield prediction \cite{Fiona2019AutomatedDO}. Methods that focus on color and shape features may overlook the rich information encoded in image textures \cite{haralick1973textural}. Therefore, there is a need to introduce novel techniques to capture the spatial heterogeneity of plant images. Lacunarity, a measure traditionally employed in fractal analysis, offers a unique perspective on texture analysis by quantifying the variability of pixel intensities within local windows or globally as shown in Figure \ref{fig:lacunarityexample}. We propose a novel texture analysis approach grounded in the concept of lacunarity \cite{Mandelbrot1967} to address the limitations of existing methods.

\begin{figure*}[htb]
\centering
	\includegraphics[width=.775\linewidth]{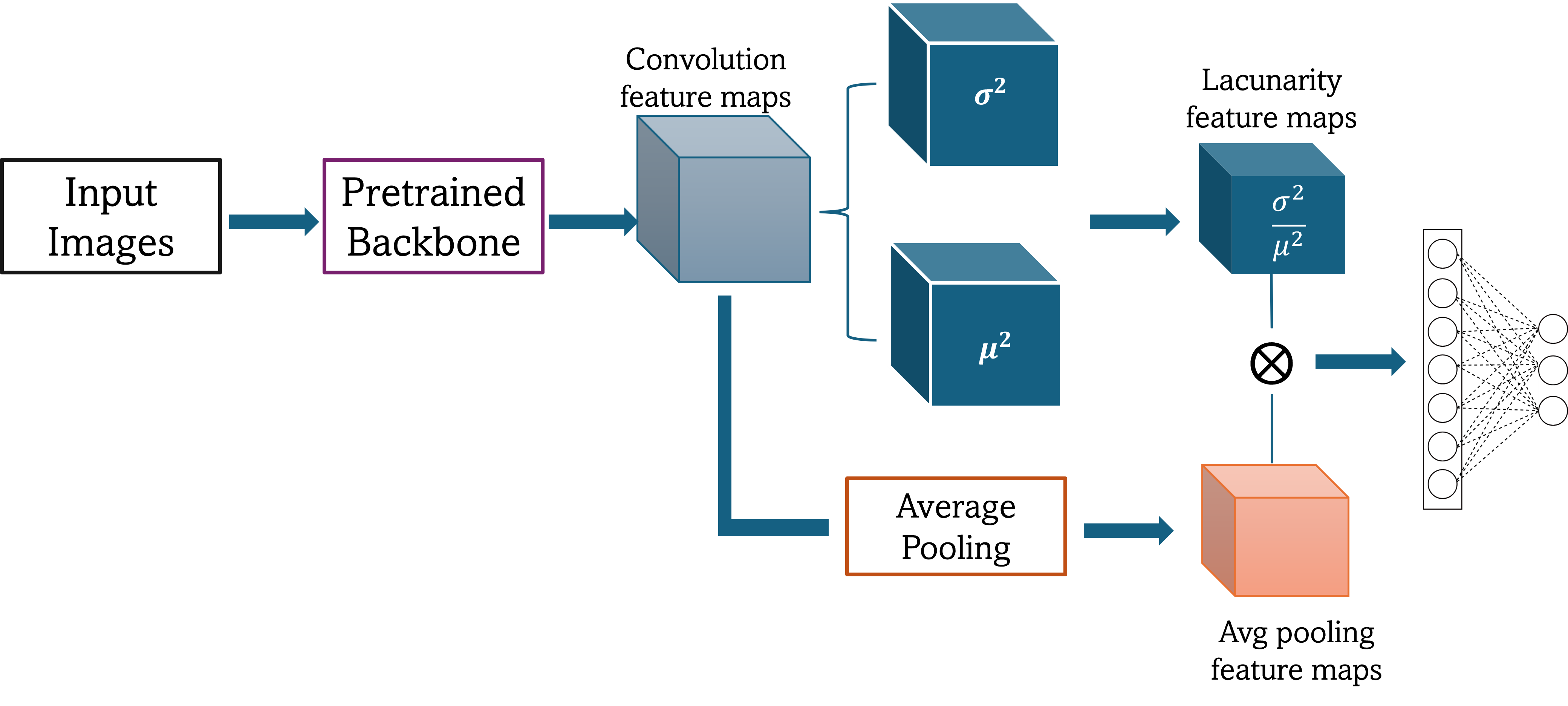}
	\caption {Overview of the proposed fusion model. Images are passed into a frozen pre-trained network and the feature maps from the final convolution layers are passed into the lacunarity pooling layer. The lacunarity features maps are then multiplied by the feature maps from the average pooling layer for enhanced analysis.}
        \label{fig:pipeline}
\end{figure*}

We propose a lacunarity pooling layer as an alternative to the conventional pooling layers. This layer leverages the spatial distribution of gap or hole sizes within images, providing a scale-dependent measure of texture. By integrating lacunarity into the network's feature extraction process, we aim to capture intricate spatial relationships for effective image classification. Our work presents several key contributions. Firstly, we introduce the lacunarity pooling layer, a novel addition to neural network architectures, which incorporates multiple methods for calculating lacunarity. Secondly, we propose several different methods to calculate lacunarity include gliding box, differential box counting, and multi-scale approaches. All of these approaches can be implemented using pre-existing layers (\textit{e.g.}, average and sum pooling). Lastly, the implementation of the lacunarity pooling layer allows for the extraction of intricate spatial features from input data, enabling more robust and versatile representations to be learned by the network. 

\section{Related Work}
\label{sec:formatting}

Lacunarity has been integrated with convolutional neural networks (CNNs) to improve feature extraction and enhance accuracy in various applications. The lacunarity cue with CNN is a novel method for saliency detection in synthetic aperture radar (SAR) images \cite{SARLacunarity}. Lacunarity identifies regions of interest (ROIs) by measuring the irregularity of scattering points within superpixels. By leveraging the irregular fluctuations in backscattering intensity, lacunarity effectively highlights ROIs in SAR imagery \cite{SARLacunarity}. Additionally, lacunarity has successfully characterized seafloor textures in synthetic aperture sonar (SAS) imagery \cite{David}.

 A related approach was formulated by Florindo \cite{florindofractalpooling} where a novel technique called fractal pooling is introduced. The conventional CNN approach for texture analysis using fully-connected layers for classification often falls short in achieving optimal performance due to the lack of invariance to rigid transforms like translation, rotation, and scaling. Global average pooling (GAP) has been traditionally employed to address this issue, but GAP results in the loss of spatial information and relationships across multiple scales. To overcome this challenge, fractal dimension is explored as an alternative pooling method for deep convolutional features in texture representation. Fractal dimension is computed over the last convolutional layer using CNN modules, summarizing higher-level features and offering a promising approach to encoding and pooling deep texture features effectively.

 Different fractal sets might have identical fractal dimensions while exhibiting diverse appearances or textures \cite{KELLER1989150}. Mandelbrot introduced the term lacunarity to describe that characteristic of fractals of the same dimension with different appearances or textures \cite{mandelbrot1982fractal}. Lacunarity serves as the complementary metric to fractal dimension, providing insight into the spatial arrangement of gaps or holes within an object. The variability or texture in gap sizes within the spatial structure of geometric objects determines the level of lacunarity. Fractal dimension measures complexity, but can give a constant value across different visual structures. Lacunarity addresses this by quantifying the ``gappiness" or heterogeneity in images, providing a more nuanced insight into their structural characteristics.

Deep Texture Recognition via Exploiting Cross-Layer Statistical Self-Similarity \cite{multiscalefractal} introduces a novel approach to texture recognition by leveraging the multi-scale nature of texture features and the self-similarity properties of textures. By analyzing the evolution of texture structures across different layers of a convolutional neural network, the proposed CLASSNet model captures multi-scale information inherent in textures. Additionally, the model exploits statistical self-similarity, a well-known property of textures characterized by fractal dimensions, to enhance the discriminative power of feature aggregation. This integration of multi-scale analysis and self-similarity fractals in texture recognition enables CLASSNet to effectively encode texture information at varying scales and improve classification accuracy. Our proposed lacunarity pooling layer also aims to incorporate multi-scale feature analysis to effectively capture texture information.

\section{Methodology}
\label{sec:litrevlacunarity}

\subsection{Lacunarity Pooling Model}
We introduce a new model, designed for integrating pre-trained CNN architectures with additional lacunarity pooling layers for improved feature extraction and classification tasks. As seen in Figure \ref{fig:pipeline}, the model takes an input image and extracts features using any pre-trained backbone (\textit{i.e.}, ConvNexT, ResNet18). For our work, the pre-trained backbone is fixed and only the final output classification layer is tuned. The backbone is fixed 1) to reduce computational costs of fine-tuning the model end-to-end and 2) to evaluate if the lacunarity pooling layer can capture features from the backbone network that are effective for classification.

The pooling layer, determined by the specified configuration, is used to spatially aggregate the feature maps. Following inspiration from \cite{florindofractalpooling}, a global average pooling (GAP) layer is employed to capture global context from the feature maps. The fusion process combines the lacunarity feature maps with the GAP feature maps, enhancing the model's ability to capture both local and global spatial information. Finally, the fused features are fed into a fully connected layer for making predictions. This flexible model architecture provides a robust framework for leveraging both pre-trained CNNs and lacunarity-based features. We investigated three approaches to compute lacunarity and each calculation is discussed in the next three subsections.

\subsection{Base Lacunarity}
\label{sec:baselacunarity}
Lacunarity is based on the pixel distribution for an image, which is obtained from scans for different box sizes at various grid orientations. Lacunarity \cite{David} can be defined through Equation \ref{eqn:base_lac}:
\begin{equation}
    L =\frac{\sigma^2}{\mu^2}
    \label{eqn:base_lac}
\end{equation}

\noindent where $\sigma^{2}$ (variance) = $\frac{1}{n} \sum_{i \in N}(x_{i} - \mu)^2$ and $\mu$ (mean) = $\frac{1}{n} \sum_{i \in N}(x_{i})$.
This formulation quantifies the heterogeneity or clustering of pixel intensities, providing a measure of lacunarity for texture analysis.

\noindent L can be therefore rewritten as shown in Equation \ref{eq:lacunarity}:
\begin{equation}
\label{eq:lacunarity}
    L = \frac{ n \Sigma_{i = 1}^{n} x_{i} ^ 2}{( \Sigma_{i = 1}^{n} x_{i}) ^2} - 1
\end{equation}

\noindent Our proposed base lacunarity pooling can be easily implemented in any deep learning framework. The numerator can simply be computed by using a sum pooling layer applied to the squared features maps multiplied by the size of the kernel ($n$). Similarly, the denominator squares the sum pooled feature maps.

By examining the distribution of gaps or holes at various scales, lacunarity provides valuable insights into the structural complexity and heterogeneity of the image. 
The process of computing lacunarity involves applying normalization to the input through the hyperbolic tangent (\text{tanh}) function to ensure that the feature values are transformed to a range between -1 and 1. To adapt the feature maps to typical pixel values ranging from 0 to 255, an additional transformation is applied as shown in Equation \ref{customtanh}:

\begin{equation}
\label{customtanh}
    x = \left(\frac{\tanh(x) + 1}{2}\right) \times 255
\end{equation}

\noindent This equation scales the output of the tanh function to the desired range for pixel values (0 to 255) commonly used in image processing. The combination of these two formulas allows for effective normalization of the input data.

\subsection{Differential Box Counting Lacunarity}
\label{sec:DBC}
The differential box counting (DBC) method is another approach for lacunarity computation \cite{Dong}. The box counting algorithm is only valid for binary images, and therefore, it is not applicable to grayscale images, as required by DBC algorithm. Briefly, DBC algorithm determines the fractal dimension value of a grayscale image by performing a least square linear fit of $\log(N_r)$ with respect to $\log(1/r)$, where $N_r$ represents the box-counting for a grid of size $r$ pixels. The box-counting process for a grid of size $r$ involves computing the difference between the boxes that denote the maximum and minimum gray levels of the pixels within the grid.

The DBC module is a novel approach to estimate lacunarity. Given input image(s) or feature map(s), the module normalizes pixel intensities and performs dilated box counting through max and min pooling operations. The lacunarity at different dilation factors \(r\) is computed using the formula shown in Equation \ref{eq:DBC}:
\begin{equation}
    L_r = \frac{{(M_r^2) \cdot Q_{mr}}}{{(M_r \cdot Q_{mr} + \epsilon)^2}}
    \label{eq:DBC}
\end{equation}

\noindent Here, \(M_r\) is the sum of \(n_r\), representing the count of boxes of size \(r\) containing pixels, and \(Q_{mr}\) is the average occupancy of these boxes. The results for various \(r\) values are concatenated and processed through a convolutional layer, enabling the module to capture heterogeneity in pixel intensities within specific regions of the image.

In the context of analyzing mass distribution in sets, Allain and Cloitre \cite{AllainandCloitre} introduced a gliding-box algorithm to define lacunarity. This method involves a box of radius $r$ sliding across a lattice superimposed on the set. Here, $n(M,r)$ represents the count of gliding-boxes with radius $r$ and mass $M$. The probability function $Q(M,r)$ is derived by normalizing $n(M,r)$ by the total box count. 
\begin{figure}[ht!]
\centering
	\includegraphics[width=.6\linewidth]{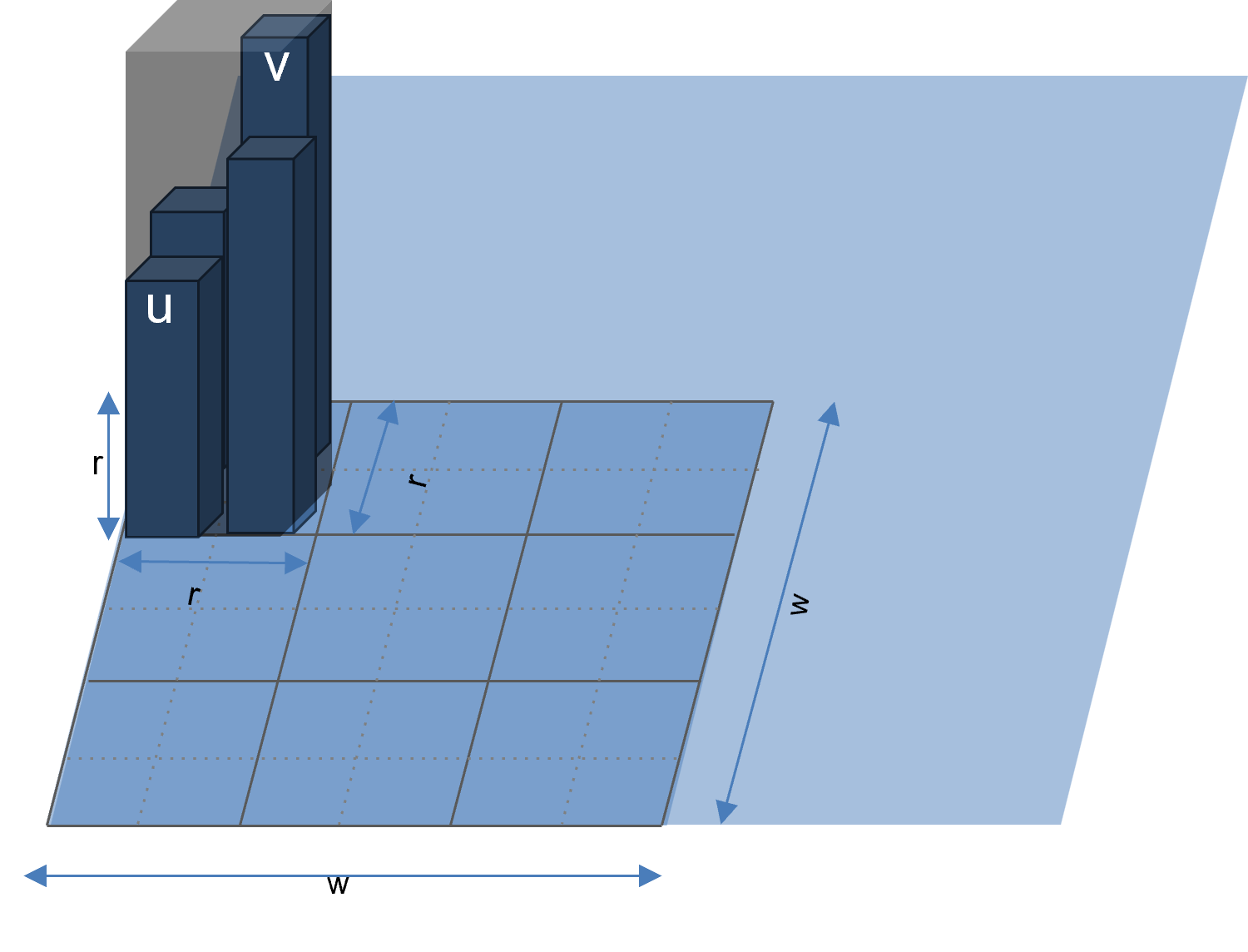}
	\caption {Illustration of differential box counting. The input image is divided into window sizes w x w, and further subdivided into r x r x r sized cubes stacked up until it covers the grayscale intensity of the local window. Here, u represents the lowest grayscale value and v represents the highest grayscale value.}
        \label{fig:dbc}
\end{figure}

The mass $M$ in Equation \ref{eq:DBC} can be computed using a DBC method introduced by Sarkar and Chaudhuri \cite{sarkar1994efficient}. As shown in Figure \ref{fig:dbc}, a cubic of size $r \times r \times r$ is positioned over the upper-left corner of an image window with dimensions $W \times W$, where $W$ is an odd number, and $r < W$. Based on pixel values, a column with more than one cubic may be required to cover the image intensity surface. Each $r$-th gliding-box is assigned numbers (\textit{e.g.}, 1, 2, 3) from bottom to top. For a given $r$-th gliding-box, with minimum and maximum pixel values falling in boxes $u$ and $v$ respectively, the relative height of the column is given by Equation \ref{eqn:height}:

\begin{equation}
n_r(i,j) = v - u - 1
\label{eqn:height}
\end{equation}

where $i$ and $j$ denote spatial coordinates. $v$ and $u$ can be computed using max and min pooling. As the $r$-th gliding-box traverses the $W \times W$ image window, the total mass is calculated through Equation \ref{eqn:mass}:

\begin{equation}
M_r = \sum_{i,j} n_r(i,j)
\label{eqn:mass}
\end{equation}

\subsection{Multi-scale Lacunarity}

\begin{figure}[ht!]
\centering
	\includegraphics[width=.6\linewidth]{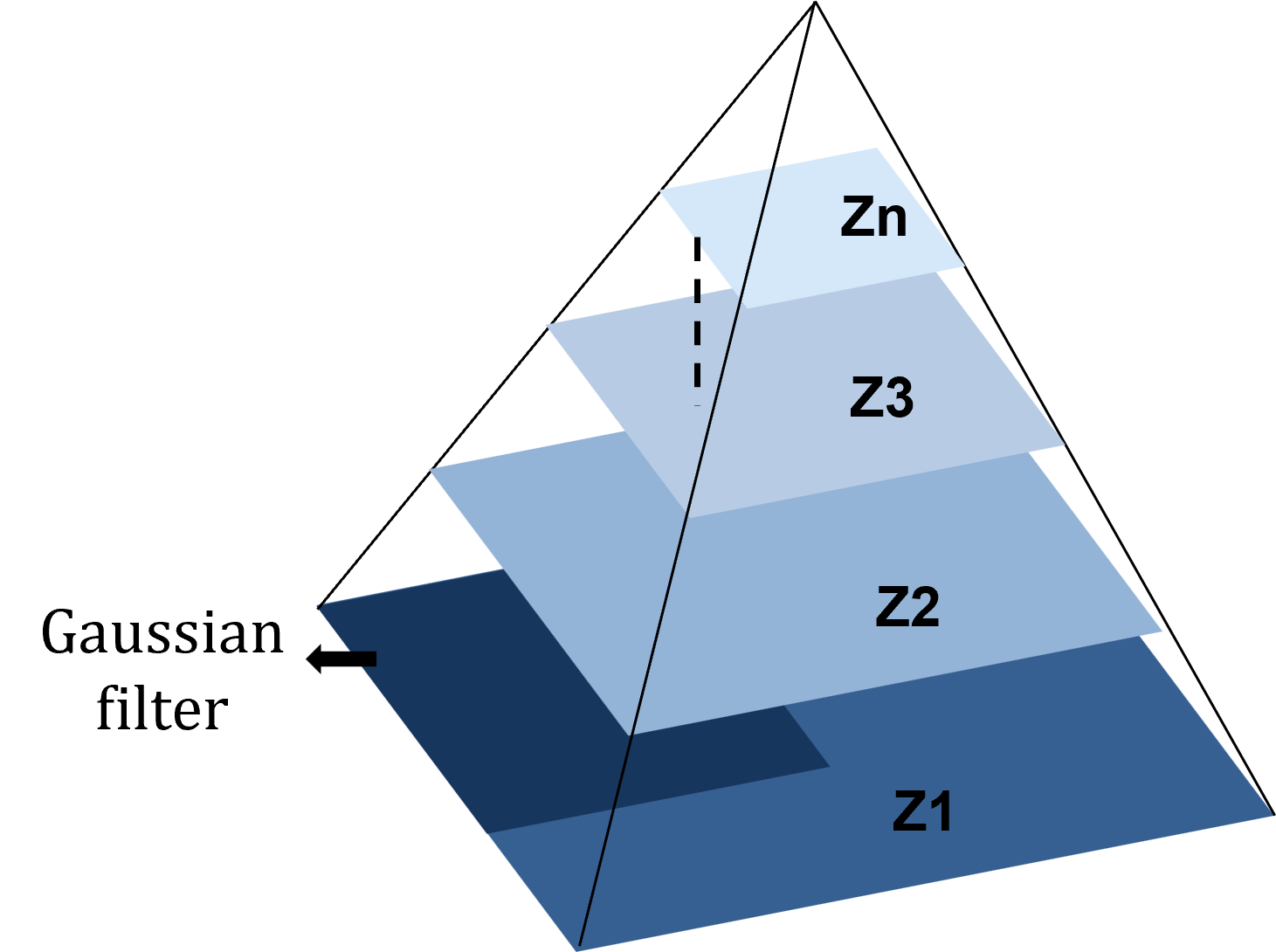}
	\caption {Multi-scale lacunarity feature approach is shown. Each input feature map is computed at different scales using a Gaussian pyramid. For each scale, the lacunarity feature map is computed and each scale is then upsampled to the same spatial resolution before concatenation.}
        \label{fig:multiscale}
\end{figure}

The multi-scale method involves computing lacunarity from different resolution levels in a Gaussian pyramid \cite{burt1987laplacian} as shown in Figure \ref{fig:multiscale}. The image undergoes recursive down-sampling and convolution with a Gaussian filter. The Kornia library \cite{riba2019kornia} is used to construct the gaussian pyramid. At each level of the pyramid, lacunarity is calculated using Equation \ref{eq:lacunarity}, capturing the spatial distribution of feature variations. Subsequently, these lacunarity values are upsampled using the bilinear interpolation to the original resolution, providing a comprehensive representation of lacunarity at different scales. The concatenated lacunarity values are then processed through a convolutional layer.

The lacunarity values are calculated at multiple scales for each channel resulting in $CS$ features maps where $C$ is the number of input channels and $S$ is the number of scales. We use a $1 \times 1$ convolution operation to perform to learn a linear combination of the different scales for each channel to ensure the number of input and output channels are the same. This will allow the layer to be be easily integrated into any existing network to replace other pooling layers.
This approach allows for a multi-scale analysis, enabling the model to capture intricate features at various levels of granularity within the image.
The multi-scale lacunarity pooling layer is summarized in Algorithm \ref{alg:multiscale}.

\begin{algorithm}[htb]
	\caption{Multi-scale Lacunarity Pooling Operation}
	\label{alg:multiscale}
	\begin{algorithmic}[1]
	\Require {$\bm{X} \in \mathbb{R}^{N \times C \times H \times W}$ feature maps, $f_{\bm{\theta}}(\bm{X})$ multi-scale operation (\textit{e.g.}, Gaussian pyramid), scale factor $S$}
        \State \begin{varwidth}[t]{\linewidth}
	    {Get scale pyramid feature maps, $\bm{Z} \gets f_
	    {\bm{\theta}}(\bm{X})$} \par
        \end{varwidth}
	\For{$s=1$ to $S$} 
	\State \begin{varwidth}[t]{\linewidth}
	    \hskip\algorithmicindent{Select scale $s$ feature maps, $\bm{Z}_{s}$} \par
		\hskip\algorithmicindent {Compute lacunarity feature maps, $\bm{L}_s$, using} \par
		\hskip\algorithmicindent {Equation \ref{eq:lacunarity}} \par
		\hskip\algorithmicindent {Upsample spatial dimension of feature maps} \par
		\hskip\algorithmicindent {based on $\bm{L}_1$ , $H' \times W'$} \par
		\hskip\algorithmicindent {Concatenate lacunarity $\bm{L}_s$ on channel}
        \par
		\hskip\algorithmicindent {dimension}
	\end{varwidth}
	\EndFor 
        \State \begin{varwidth}[t]{\linewidth}
	    {Use $1 \times 1$ convolution to reduce $\bm{L} \in \mathbb{R}^{N \times CS \times H' \times W'}$ to $\bm{L} \in \mathbb{R}^{N \times C \times H' \times W'}$} to combine lacunarity across different scales\par
        \end{varwidth}
	\State {\Return $\bm{L}$}
	\end{algorithmic}
\end{algorithm}

\section{Experimental Results}

\subsection{Experimental Setup}
The following experimental setup was used for training: a batch size of 128, cross-entropy loss function, Adam optimization with an initial learning rate of $0.001$, a total of 100 epochs, and early stopping implemented after 10 epochs if the validation loss did not improve.
The experiments were conducted on two NVIDIA A100 GPUs. The datasets were split into 70$\%$ training, 10$\%$ validation, and 20$\%$ testing sets. We evaluated the performance of three CNN models: ConvNeXt \cite{liu2022convnet}, ResNet18 \cite{resnet18}, and DenseNet161 \cite{huang2018densely}, using various pooling strategies for image classification. A similar data augmentation procedure to \cite{Mohan_2023,xue2018deep,peeples2021histogram} was added to the training data: random cropping and horizontal flipping. In the implementation of the multi-scale lacunarity pooling layer, we use two scales ($S = 2$). The scale was set to this value as the spatial size of the feature maps into the pooling layer were $7 \times 7$.

\begin{table*}[htb]
\centering
\caption[Accuracy comparison of PlantVillage]{Average test accuracy with $\pm$ 1 standard deviation of all pooling method for PlantVillage across three experimental trials. The best average accuracy is bolded for the pooling layers.}
\label{Table accuracy PlantVillage}
\begin{tabular}{|c|c|c|c|c|ccc|}
\hline
Models      & Max        & Average    & L$_2$         & Fractal    & \multicolumn{3}{c|}{Lacunarity Pooling (Ours)}                                  \\ \hline
            &            &            &            &            & \multicolumn{1}{c|}{Base}       & \multicolumn{1}{c|}{DBC}        & Multi-scale \\ \hline
ConvNeXt & 91.71±0.26 & 97.30±0.39 & 97.30±0.39 & 95.74±0.54      & \multicolumn{1}{c|}{96.48±0.12}  & \multicolumn{1}{c|}{\textbf{97.71±0.17}} & 97.49±0.30 \\ \hline
Resnet18 & 91.50±1.26  & 95.57±0.21 & 95.57±0.17 & 90.68±0.50     & \multicolumn{1}{c|}{95.32±0.44}  & \multicolumn{1}{c|}{\textbf{96.15±0.20}} & 95.35±0.24 \\ \hline
Denesenet161 & 95.47±0.53 & 97.49±0.04 & 96.32±0.17 & 98.01±0.11 & \multicolumn{1}{c|}{97.66±0.13} & \multicolumn{1}{c|}{97.75±0.11} & \textbf{98.07±0.21} \\ \hline
\end{tabular}
\end{table*}

\begin{table*}[htb]
\centering
\caption[Accuracy comparison of LeavesTex1200]{Average test accuracy with $\pm$ 1 standard deviation of all pooling method for LeavesTex1200 across three experimental trials. The best average accuracy is bolded for the pooling layers.}
\label{Table accuracy LeavesTex}
\centering
\begin{tabular}{|c|c|c|c|c|ccc|}
\hline
Models      & Max        & Average    & L$_2$         & Fractal    & \multicolumn{3}{c|}{Lacunarity Pooling (Ours)}                                  \\ \hline
            &            &            &            &            & \multicolumn{1}{c|}{Base}       & \multicolumn{1}{c|}{DBC}        & Multi-scale \\ \hline
ConvNeXt    & 86.88±1.81 & 94.88±1.23 & 94.88±1.23 & 94.44±0.87 & \multicolumn{1}{c|}{83.22±2.35} & \multicolumn{1}{c|}{89.22±1.28} & \textbf{95.00±0.82}     \\ \hline
ResNet18    & 84.44±1.09 & 88.89±0.31 & 88.89±0.31 & 88.22±0.16 & \multicolumn{1}{c|}{86.11±0.87} & \multicolumn{1}{c|}{87.67±2.32} & \textbf{90.11±2.79}  \\ \hline
DenseNet161 & 86.33±1.78 & 93.44±1.13 & 89.78±1.85 & 94.11±1.97 & \multicolumn{1}{c|}{91.11±0.31} & \multicolumn{1}{c|}{93.22±2.04}          & \textbf{94.55±0.31}           \\ \hline
\end{tabular}
\end{table*}

\begin{table*}[h!]
\caption[Accuracy comparison of DeepWeeds]{Average test accuracy with $\pm$ 1 standard deviation of all pooling method for DeepWeeds across three experimental trials. The best average accuracy is bolded for the pooling layers.}
\label{Table accuracy DeepWeeds}
\centering
\begin{tabular}{|c|c|c|c|c|ccc|}
\hline
Models      & Max        & Average    & L$_2$         & Fractal    & \multicolumn{3}{c|}{Lacunarity Pooling (Ours)}                     \\ \hline
            &            &            &            &            & \multicolumn{1}{c|}{Base} & \multicolumn{1}{c|}{DBC} & Multi-scale \\ \hline
ConvNeXt    & 70.83±0.71 & 85.53±0.78 & 85.53±0.78 & 85.37±0.29          & \multicolumn{1}{c|}{80.42±0.57}    & \multicolumn{1}{c|}{83.16±0.20}   & \textbf{86.01±0.26}           \\ \hline
ResNet18    & 66.59±1.52 & 73.27±0.48 & 73.27±0.48 & 63.68±2.53 & \multicolumn{1}{c|}{69.95±0.40}    & \multicolumn{1}{c|}{69.95±0.75}   & \textbf{73.77±0.03}           \\ \hline
DenseNet161 & 74.94±0.75 & 79.66±1.09 & 75.64±0.42 & \textbf{85.69±0.40}          & \multicolumn{1}{c|}{74.30±0.20}    & \multicolumn{1}{c|}{75.29±0.90}   & 80.44±0.32           \\ \hline
\end{tabular}
\end{table*}

\subsection{Benchmark Datasets Results}

\begin{figure*}[h!]
    \centering
    \setlength{\fboxsep}{0pt} 
    \begin{subfigure}[b]{0.17\linewidth} 
        \centering
        \includegraphics[width=\linewidth]{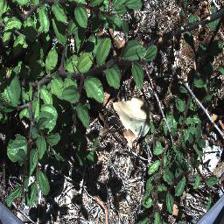}
        \caption{Chinee Apple}
        \label{fig:snakeweeds}
    \end{subfigure}\hfill 
    \begin{subfigure}[b]{0.17\linewidth} 
        \centering
        \includegraphics[width=\linewidth]{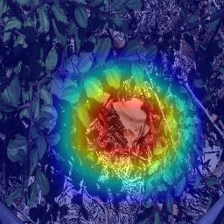}
        \caption{Average}
        \label{avgpool}
    \end{subfigure}\hfill 
    \begin{subfigure}[b]{0.17\linewidth}
        \centering
        \includegraphics[width=\linewidth]{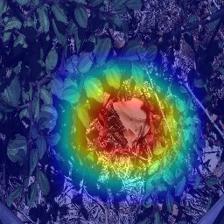}
        \caption{Max}
        \label{maxpool}
    \end{subfigure}\hfill
    \begin{subfigure}[b]{0.17\linewidth}
        \centering
            \includegraphics[width=\linewidth]{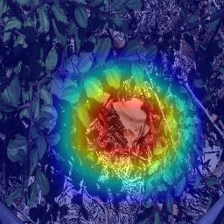}
        \caption{L$_2$}
        \label{l2pool}
    \end{subfigure}\hfill
    \begin{subfigure}[b]{0.17\linewidth}
        \centering
        \includegraphics[width=\linewidth]{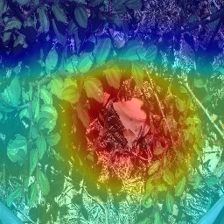}
        \caption{MS Lacunarity (Ours)}
        \label{lacunarity}
    \end{subfigure}\hfill
\caption{Example results of DeepWeeds dataset across pooling layers for the Resnet18 model. EigenCAM \cite{muhammad2020eigen} was used to create the class activation maps to show which areas of the image each model focused on. As we can see in this example, the multi-scale lacunarity pooling layer captures useful information across varying spatial levels of the image.}
\end{figure*}

\begin{figure*}[htb]
    \centering
	\begin{subfigure}{.16\textwidth}{
			\includegraphics[width=\textwidth]{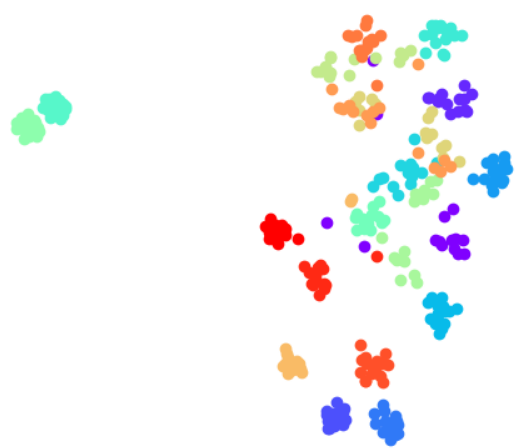}
			\caption{Avg \\ \centering 38.80 $\pm$ 1.26}
			\label{fig:avg}
		}
	\end{subfigure}
	\centering
	\begin{subfigure}{.16\textwidth}{
	\includegraphics[width=\textwidth]{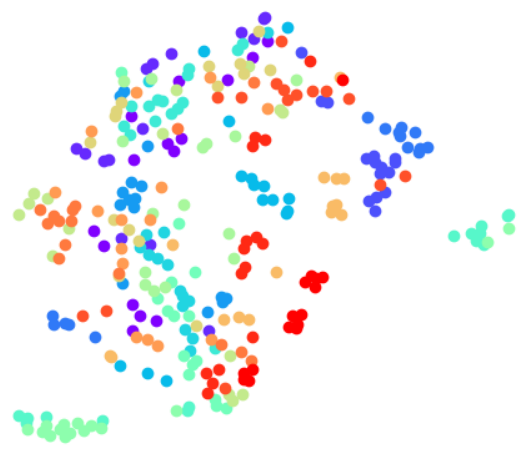}
			\caption{Max \\ \centering 35.05 $\pm$ 1.09}
			\label{fig:max}
		}
	\end{subfigure}
	\centering
	\begin{subfigure}{.16\textwidth}{
	\includegraphics[width=\textwidth]{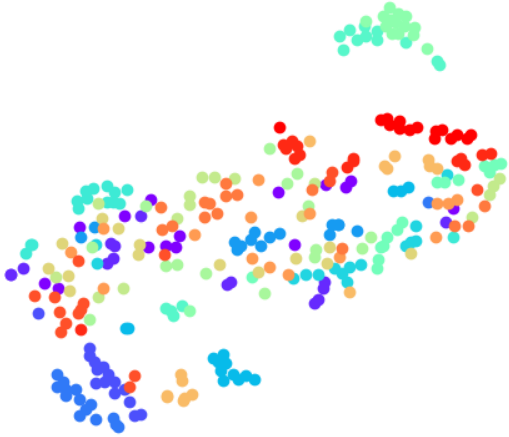}
			\caption{L$_2$ \\ \centering 35.67 $\pm$ 1.20}
			\label{fig:L2}
		}
	\end{subfigure} 
	\centering
	\begin{subfigure}{.16\textwidth}{
	\includegraphics[width=\textwidth]{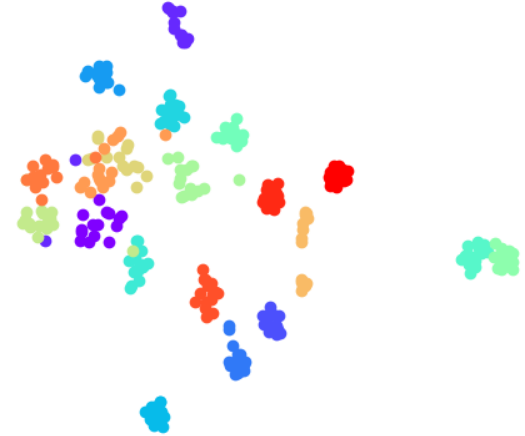}
			\caption{Fractal \\ \centering 38.58 $\pm$ 1.28}
			\label{fig:Fractal_embed}
		}
	\end{subfigure}
	\centering
		\begin{subfigure}{.16\textwidth}{
	\includegraphics[width=\textwidth]{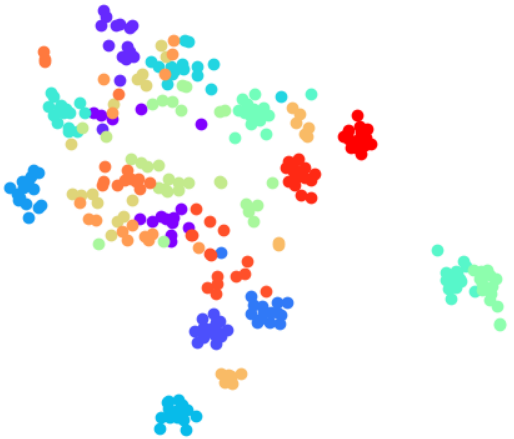}
			\caption{DBC Lacunarity\\  \centering 38.46 $\pm$ 1.91}
			\label{fig:MS_embed}
		}
	\end{subfigure}
	\centering
	\begin{subfigure}{.16\textwidth}{
	\includegraphics[width=\textwidth]{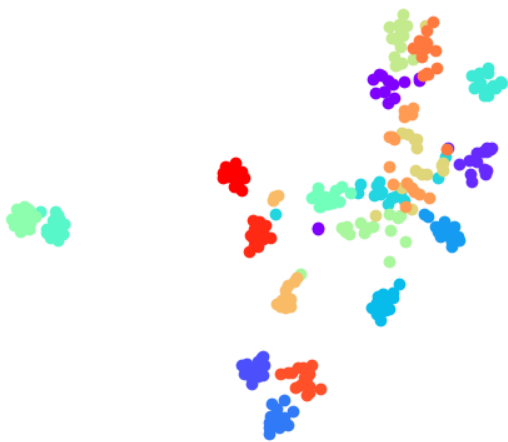}
			\caption{MS Lacunarity \\ \centering \textbf{39.25 $\pm$ 1.39}}
			\label{fig:Lac_MS}
		}
	\end{subfigure}
	\caption{t-SNE results for LeavesTex1200 Dataset trained on the Densenet161 model. The colors represent the 20 different classes from LeavesTex1200 dataset. Each t-SNE used the same random seed so there is a fair comparison between the different pooling layers. We also computed the average log Fisher Discriminant Ratio ($\pm 1$ standard deviation) using the features from the penultimate layer. Higher scores indicate more compact and well separated classes in the higher dimensional space. Multi-scale (MS) lacunarity is shown in Figure \ref{fig:Lac_MS}. Our quantitative metrics match the qualitative observations in the t-SNE plot.}
	\label{fig:Embeddings}
 \end{figure*}

\begin{figure*}[htb]
    \centering
	\begin{subfigure}{.494\textwidth}{
			\includegraphics[width=\textwidth]{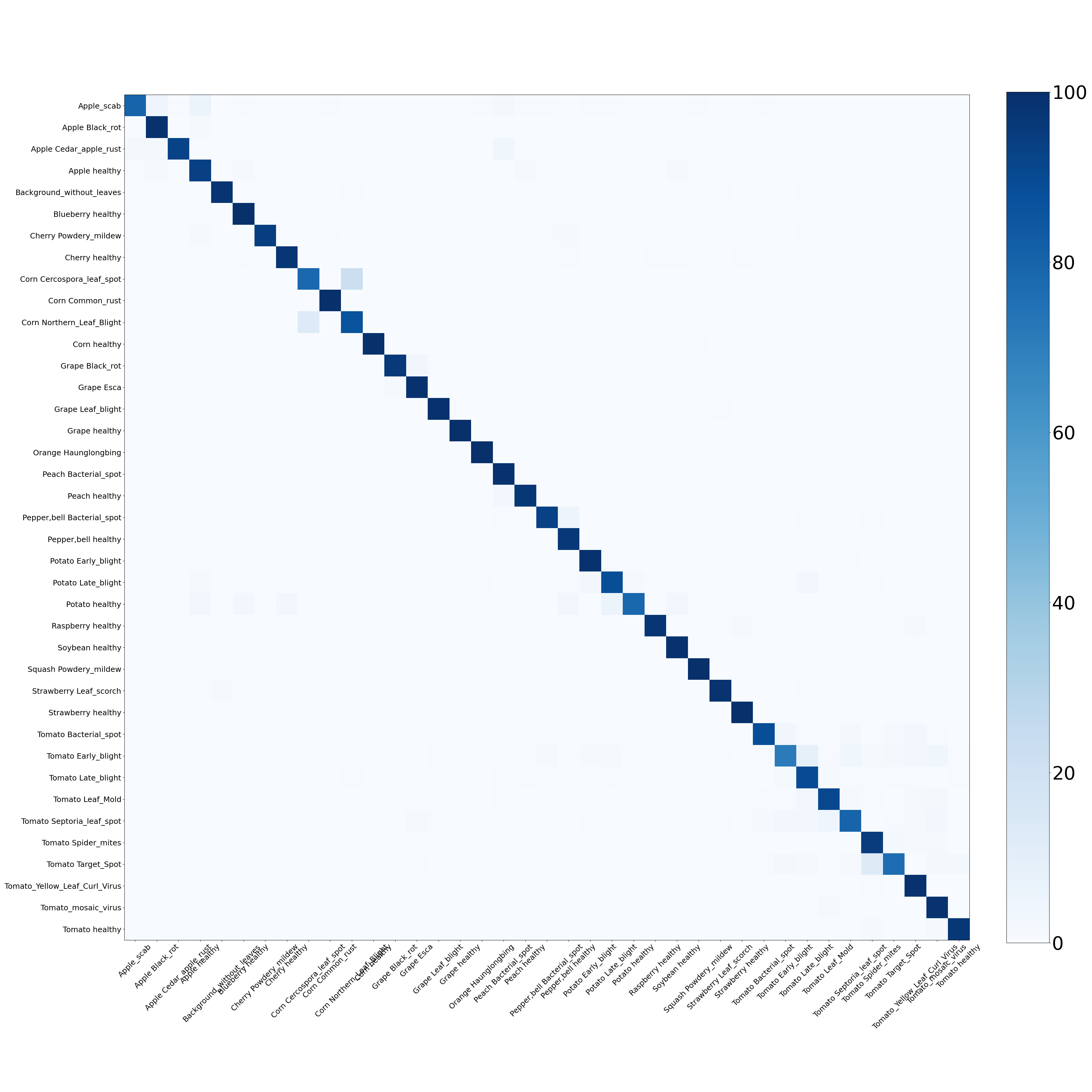}
        \caption{Fractal pooling \\ 95.74 $\pm$ 0.54
        }
        \label{fig:fractal_cm}
		}
	\end{subfigure}
	\centering
	\begin{subfigure}{.494\textwidth}{
	\includegraphics[width=\textwidth]{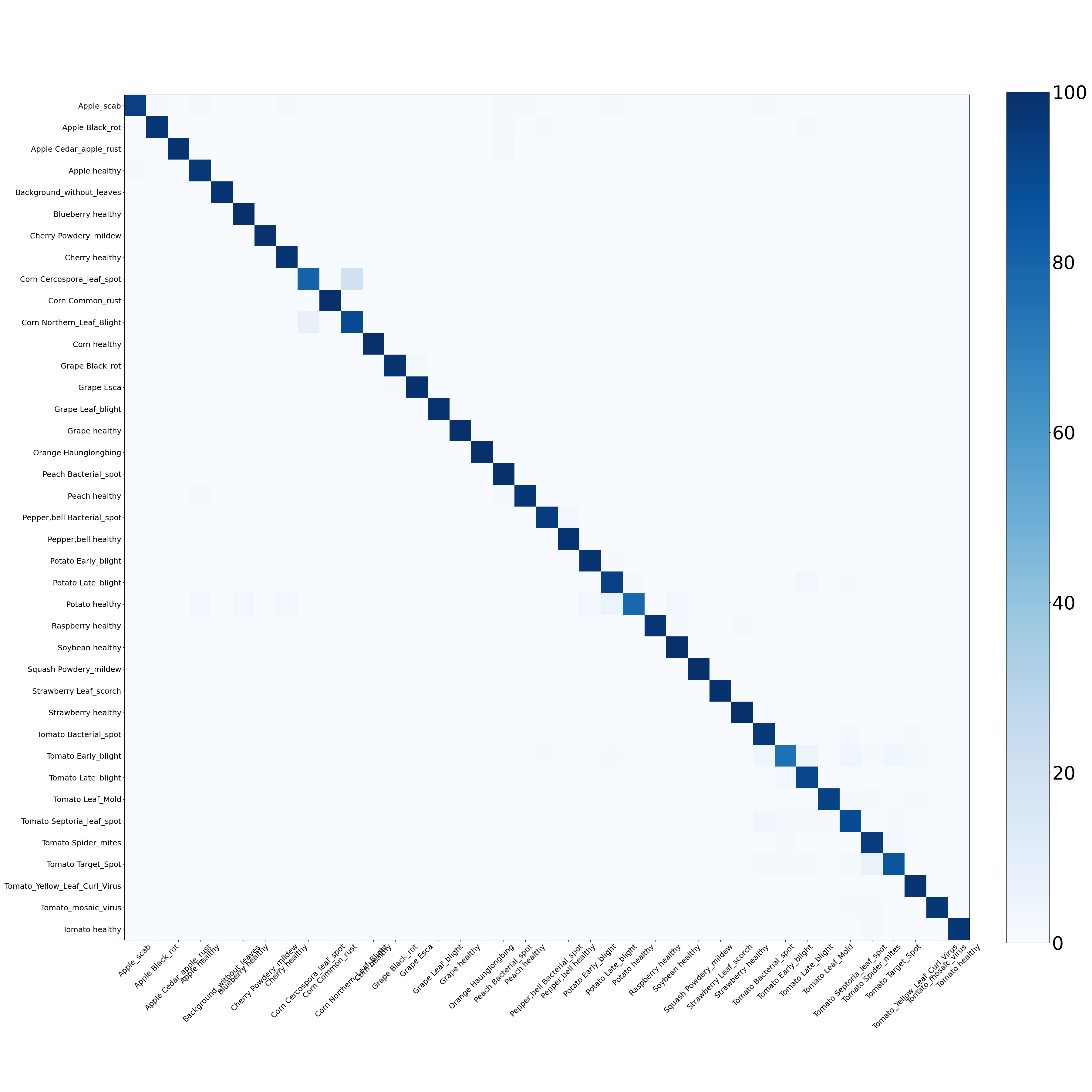}
			\caption{DBC Lacunarity Pooling \\ \textbf{97.71 $\pm$ 0.17}}
			\label{fig:lac_cm}
		}
	\end{subfigure}
        
 \caption{Confusion matrices of PlantVillage dataset for fractal (\ref{fig:fractal_cm}) and DBC lacunarity (\ref{fig:lac_cm}) pooling layers in the ConvNeXt model respectively. Lacunarity improved performance for most classes (34 of 39) demonstrating the effectiveness of our proposed lacunarity pooling layer in comparison to fractal pooling. The average classification accuracy with $\pm$1 standard deviation is shown below each confusion matrix.}
 \label{fig:confusion matrices}
 \end{figure*}

\textbf{PlantVillage}
The PlantVillage dataset \cite{hughes2015open}
is for multi-class image classification with
55,448 images divided into
39 classes representing background-only (out of domain images such as animals and buildings), healthy, and diseased plants. Images span 14 plant species: Apple, Blueberry, Cherry, Corn, Grape, Orange, Peach, Bell Pepper, Potato, Raspberry, Soybean, Squash, Strawberry, and Tomato and contains images of 17 fungal diseases, four bacterial diseases, two mold (Oomycete) diseases, two viral diseases, and one disease caused by a mite.

As seen in Table \ref{Table accuracy PlantVillage}, the DenseNet161 model consistently outperformed the other two models, achieving the highest classification accuracy across all pooling methods. Specifically, DenseNet161 with GAP yielded an average accuracy of 97.49$\%$, while our pooling layer yields an improvement in accuracy to 98.07$\%$. Max pooling selects the highest values within the local region. This can be problematic when dealing with images that have varying lighting conditions or shadows such as leaf images since max pooling is sensitive to outliers \cite{peeples2021histogram}. In such cases, these extreme values from max pooling might not represent the overall characteristics of the leaf accurately. On the other hand, average pooling or the other pooling layers used calculate ``smoothed" values giving a more balanced representation across the local regions and being less influenced by extreme values. The recently proposed fractal pooling technique \cite{florindofractalpooling}, which aims to capture the fractal nature of textures, has a comparable classification accuracy in our experiments for DenseNet161 and ConvNeXt in comparison to the baseline GAP models, but this comes at a computational cost as discussed in Section \ref{sec:params}. The DBC method shows an improvement in the accuracy for the ResNet18 and ConvNeXt models without adding additional learnable parameters. 

\textbf{LeavesTex1200}
The real-world application, named LeavesTex1200 \cite{casanova2009plant}, employs commercial scanner-captured leaf surface images to classify Brazilian plant species. The dataset consists of 20 species, each represented by 60 images, totaling 1200 samples. Specimens were collected in vivo, cleaned, and aligned along the basal/apical axis. From each specimen, three non-overlapping 128 × 128 pixel windows were extracted to create texture images for classification purposes.
As seen in Table \ref{Table accuracy LeavesTex} the multi-scale lacunarity performs the best when compared to our base lacunarity method. LeavesTex1200 images are distinctly focused on leaves without major variations in texture across a single scale. The leaves have a homogeneous vein structure in the leaves. This lack of textural heterogeneity makes the base lacunarity less effective since the base lacunarity is optimized for capturing textural variations within a single scale. However, the multi-scale lacunarity pooling layer becomes more adept at capturing subtle texture changes across different scales. When zoomed out, leaves exhibit greater textural heterogeneity, which the multi-scale method effectively leverages, resulting in improved performance compared to both the base lacunarity and baseline average pooling layers.

\textbf{DeepWeeds}
DeepWeeds \cite{DeepWeeds2019} comprises a total of 17,509 RGB images. These images were captured by a specialized ground weed control robot under natural field conditions, without controlled lighting. The dataset covers eight distinct weed species along with various non-weed plants for a total of nine classes. Each weed species is represented by more than 1000 images.
DeepWeeds demonstrates better accuracy with the lacunarity pooling layer across ConvNeXt and ResNet18 models as shown in Table \ref{Table accuracy DeepWeeds}. However, notable accuracy improvements are particularly evident with the ConvNeXt model. Analysis of the results shows that fractal pooling exhibits better performance, especially with deeper models such as DenseNet161 \cite{florindofractalpooling}. Fractal pooling also uses residual block to improve the feature representation captured by this layer. DenseNet161 uses feature concatenation across multiple layers that enables the integration of features representing various levels of abstraction. In texture images, these features correspond to patterns observed at different scales. This multi-scale analysis has long been recognized as crucial for texture analysis \cite{coburn2004multiscale}, so it is expected that DenseNet models achieves the best accuracy in this context.

Comparing results across all the pooling layers in Tables \ref{Table accuracy PlantVillage}, \ref{Table accuracy LeavesTex}, and \ref{Table accuracy DeepWeeds}, it is evident that the lacunarity pooling layers consistently outperform the baseline average pooling layers on average. Specifically, the multi-scale approach demonstrates better performance compared to the base method described in Section \ref{sec:baselacunarity}. The DBC method, as discussed in Section \ref{sec:DBC}, performs comparably to the average pooling layer. 
The operations involved in average pooling is similar with L$_2$ pooling, which is a form of power-average pooling. Given that we focus on feature extraction in our experiments, it is observed that the results for the L$_2$ pooling closely resemble those obtained through average pooling.

The reason why our lacunarity pooling method, especially the multi-scale variant, outperforms others lies in its capability for capturing the intricate details and varying complexities present in our image data. Unlike typical average pooling techniques that might overlook subtle nuances or textural variations, our multi-scale lacunarity pooling delves into different levels of detail, thereby increasing the model's ability to discern patterns. Moreover, our pooling approach is finely tuned to use the self-similarity and spatial irregularities inherent in natural textures, which are prevalent in the wide array of plant images we dealt with across different datasets. 

\subsection{Explainable AI Analysis}
We used an Explainable AI technique, EigenCAM, \cite{muhammad2020eigen} to visualize the salient regions identified in an input image by the lacunarity pooling layer and other pooling approaches with the ResNet18 model. EigenCAM showed more pronounced areas of significance in the input image, characterized by wider regions of activation for our proposed lacunarity pooling layer fusion model. The higher regions of interest, represented by the brighter highlighted areas, signifies stronger and more focused attention on crucial features within the images. This emphasis on specific regions is advantageous as this aids in capturing intricate details across varying spatial levels of the image. Compared to the other pooling layers such as max, average, and L$_2$, we observed improved performance with our model in terms of capturing more information across the input image. This suggests that our proposed lacunarity pooling strategy captures more extensive spatial information, thereby potentially enhancing the discriminative power of the model's predictions.

\subsection{t-SNE Visualization Analysis}
The t-Distributed Stochastic Neighbor Embedding (t-SNE) \cite{TSNE} was used to visualize the high-dimensional feature representations learned by our fusion model and other pooling layers. The same random seed was used to fairly compare each t-SNE projection of the features extracted from each pooling approach. For max pooling as shown in Figure \ref{fig:max}, there is not much separation or compactness in the projected space (similar to L$_2$ in Figure \ref{fig:L2}). However, the lacunarity pooling layers appear to form more compact and separable clusters, especially the multi-scale lacunarity approach.

Moreover, to quantitatively assess the discriminative power of features extracted by our fusion model, we computed the log Fisher Discriminant Ratio (FDR) \cite{peeples2021histogram} scores for each class in the original feature space. The FDR score measures the ratio of between-class variation to within-class variation, with higher values indicating greater discriminatory ability. In our experiments, a higher FDR score signifies that the features are more effective at differentiating between classes, thus contributing to enhanced classification performance. When examining the FDR scores across different pooling layers in Figure \ref{fig:Embeddings}, we observed notable differences. Particularly, the multi-scale lacunarity pooling layer consistently yielded higher FDR scores compared to the DBC lacunarity pooling layer. This suggests that the multi-scale variant of the lacunarity pooling layer captures more discriminative information from the input data, leading to better separability between classes.

\subsection{Confusion Matrix Analysis}
When comparing the confusion matrices generated by different pooling methods for PlantVillage, we observed improvements in the classification performance of our lacunarity pooling multi-scale approach across a majority of classes. Specifically, out of the total 39 classes in our dataset, our method demonstrated better classification accuracy in 34 classes compared to the fractal pooling layer \cite{florindofractalpooling} as seen in Figure \ref{fig:confusion matrices}. For classes like Apple Scab the accuracy increased from 79.84$\%$ to 93.65$\%$ and Tomato early blight went up from 70.85$\%$ to 80.61$\%$. Classes which were misclassified as Corn Cercospora leaf spot instead of Corn Northern leaf blight, where the false positive percentage at 21.50$\%$ was reduced to 19.63$\%$. This indicates an enhancement in the model's ability to correctly classify a wide array of plant diseases and species. The confusion matrices vividly illustrate how our lacunarity pooling method effectively captures the intricate features and subtle variations present in the images, leading to more accurate predictions across diverse classes.

\subsection{Model Parameter Analysis}
\label{sec:params}
\begin{table}[htb]
\centering
\caption{Comparison of learnable parameters introduced by fractal and multi-scale (MS) lacunarity pooling layers. All other pooling layers (average, max, L$_2$, base lacunarity, and DBC lacunarity) do not introduce additional parameters to the models. Our proposed MS lacunarity pooling layer learns information across multiple levels of the image while being more computationally efficient than the fractal pooling layer.}
\label{Table LearnableParams}
\begin{tabular}{|c|c|c|}
\hline
Model       & Fractal & MS Lacunarity \\ \hline
ResNet18    & 268,297         & 1,536                 \\ \hline
ConvNeXt    & 599,049         & 2,304                 \\ \hline
DenseNet161 & 4,901,769       & 6,624                 \\ \hline
\end{tabular}
\end{table}

Table \ref{Table LearnableParams} outlines the number of learnable parameters for each model, highlighting the parameter increase introduced by the fractal method and multi-scale lacunarity across all models. Despite the additional parameters introduced by the fractal method, there is no statistically significant improvement in accuracy compared to the baseline layers except for the DeepWeeds dataset. The multi-scale lacunarity pooling layer introduces extra parameters due to the $1 \times 1$ convolution; however, there are improvements in the model's performance. In the case of DBC for PlantVillage, the DBC lacunarity was a viable alternative to the GAP baseline models to achieve a slight increase in average overall performance without additional learnable parameters. 

\section{Conclusion}
The proposed lacunarity pooling layer is as a novel approach to enhance the performance of CNNs in computer vision tasks.  We presented a new model that integrates lacunarity pooling layers with existing CNN architectures, facilitating the extraction of both local and global spatial information. Through extensive experimentation and analysis, we have demonstrated the efficacy of our approach across multiple datasets and model architectures. Our findings reveal that incorporating lacunarity pooling layers can lead to improved feature extraction and classification accuracy, particularly evident in models such as DenseNet161. 

While certain variations in pooling methods gave lower accuracy, our study shows the adaptability and potential of lacunarity-based features in enhancing CNN performance. Future work includes extending this pooling layer to other data modalities (\textit{e.g.}, hyperspectral data), investigating other backbone architectures and adapting the lacunarity pooling layer for other tasks such as segmentation. Another interesting direction is to integrate the multi-scale lacunarity layer with features from different levels of the network. Ultimately, our unique pooling layer demonstrates the potential for the integration of novel textural features like lacunarity as valuable additions to CNN architectures.

\section{Acknowledgement} Portions of this research were conducted with the advanced computing resources provided by Texas A\&M High Performance Research Computing.

{
    \small
    \bibliographystyle{ieeenat_fullname}
    \bibliography{main}
}


\end{document}